\documentclass[conference]{IEEEtran}
\IEEEoverridecommandlockouts
\usepackage{cite}
\usepackage{amsmath,amssymb,amsfonts}
\usepackage{algorithmic}
\usepackage{graphicx}
\usepackage{textcomp}
\usepackage{xcolor}
\usepackage{comment}
\def\BibTeX{{\rm B\kern-.05em{\sc i\kern-.025em b}\kern-.08em
    T\kern-.1667em\lower.7ex\hbox{E}\kern-.125emX}}
\begin{document}

\title{Adaptive tracking control for task-based robot trajectory planning\\
\thanks{This work was supported by UARE program to L. Trucios}
}

\author{\IEEEauthorblockN{Luis E. Trucios}
\IEEEauthorblockA{\textit{Department of Mechanics} \\
\textit{Universidad Nacional de Ingenieria}\\
Lima, Peru \\
luis.trucios.r@uni.pe}
\and
\IEEEauthorblockN{Mahdi Tavakoli}
\IEEEauthorblockA{\textit{Faculty of Engineering} \\
\textit{University of Alberta}\\
Edmonton, Canada \\
mahdi.tavakoli@ualberta.ca}
\and
\IEEEauthorblockN{Kim Adams}
\IEEEauthorblockA{\textit{Faculty of Rehabilitation Medicine} \\
\textit{University of Alberta}\\
Edmonton, Canada \\
kdadams@ualberta.ca}
}

\maketitle

\begin{abstract}
This paper presents a “Learning from Demonstration” method to perform robot movement trajectories that can be defined as you go. This way unstructured tasks can be performed, without the need to know exactly all the tasks and start and end positions beforehand. 
The long-term goal is for children with disabilities to be able to control a robot to manipulate toys in a play environment, and for a helper to demonstrate the desired trajectories as the play tasks change. A relatively inexpensive 3-DOF haptic device made by Novint is used to perform tasks where trajectories of the end-effector are demonstrated and reproduced.
Under the condition where the end-effector carries different loads, conventional control systems possess the potential issue that they cannot compensate for the load variation effect.  Adaptive tracking control can handle the above issue. Using the Lyapunov stability theory, a set of update laws are derived to give closed-loop stability with proper tracking performance.
\end{abstract}


\section{Introduction}
Children who have severe disabilities have difficulty playing with toys because of limitations reaching out, holding, and manipulating toys.  Robots are a potential tool to provide manipulation because children can control them with residual functional abilities (i.e., switches, brain signals), but it is important that play tasks can be changed quickly to keep a child's interest.  For example, the pickup and drop-off locations of a toy should be easily reprogrammed, and the type of toy supported by the robot should be easily replaced.   Instead of reprogramming for every location and toy weight, a child's playmate could show the robot the new task requirements.

The problem of learning and imitating a task through observation is inherently easy for humans, but a robot has to be re-programmed to perform a different task \cite{b1}. Learning from Demonstration (LfD), also known as imitation learning, provides an intuitive way to readily teach new trajectories to a robot.  Instead of re-programming every new task, a task is learned from demonstrations \cite{b2}.  This means a person can move the end-effector of the robot along the desired trajectory, and the robot can learn that trajectory so that it can replay it independently of the person. In this way, a child who has a physical disability can instigate the robot trajectory with whatever action they can (e.g., physical movement at a control interface or a physiological signal) to manipulate toys independently with the robot.  When they want to do a new play task, a helper can demonstrate the new trajectory to the robot. 

As the haptic robot dynamics and the environment contain various uncertainties and disturbances in practical applications, one of the effective ways to deal with this difficulty is to apply an adaptive control. In  \cite{b3}, \cite{b4} and \cite{b5} a model reference adaptive control (MRAC) for a constrained manipulator is applied to reduce model uncertainties.
Since joint acceleration is difficult to measure precisely, in \cite{b6} an adaptive impedance controller is designed without requiring estimates of acceleration.

For the purpose of having higher repeatability and accuracy, an adaptive control system is able to adjust parameters related to the robot model when the end-effector load changes frequently. In \cite{b7} a summary about "MRAC" shows that traditional controllers are not able to handle systems where their dynamic model parameters varies with respect to time.

This paper aims to design a control algorithm to perform demonstrated tasks online. This algorithm is better at handling the significant non-linearities in the system and different weight of toys that usually makes control and estimation tasks for this class of robots difficult using classical methods \cite{b8}.

\section{Experimental Setup}
\begin{figure*}[htbp]
\centerline{\includegraphics[scale=0.15]{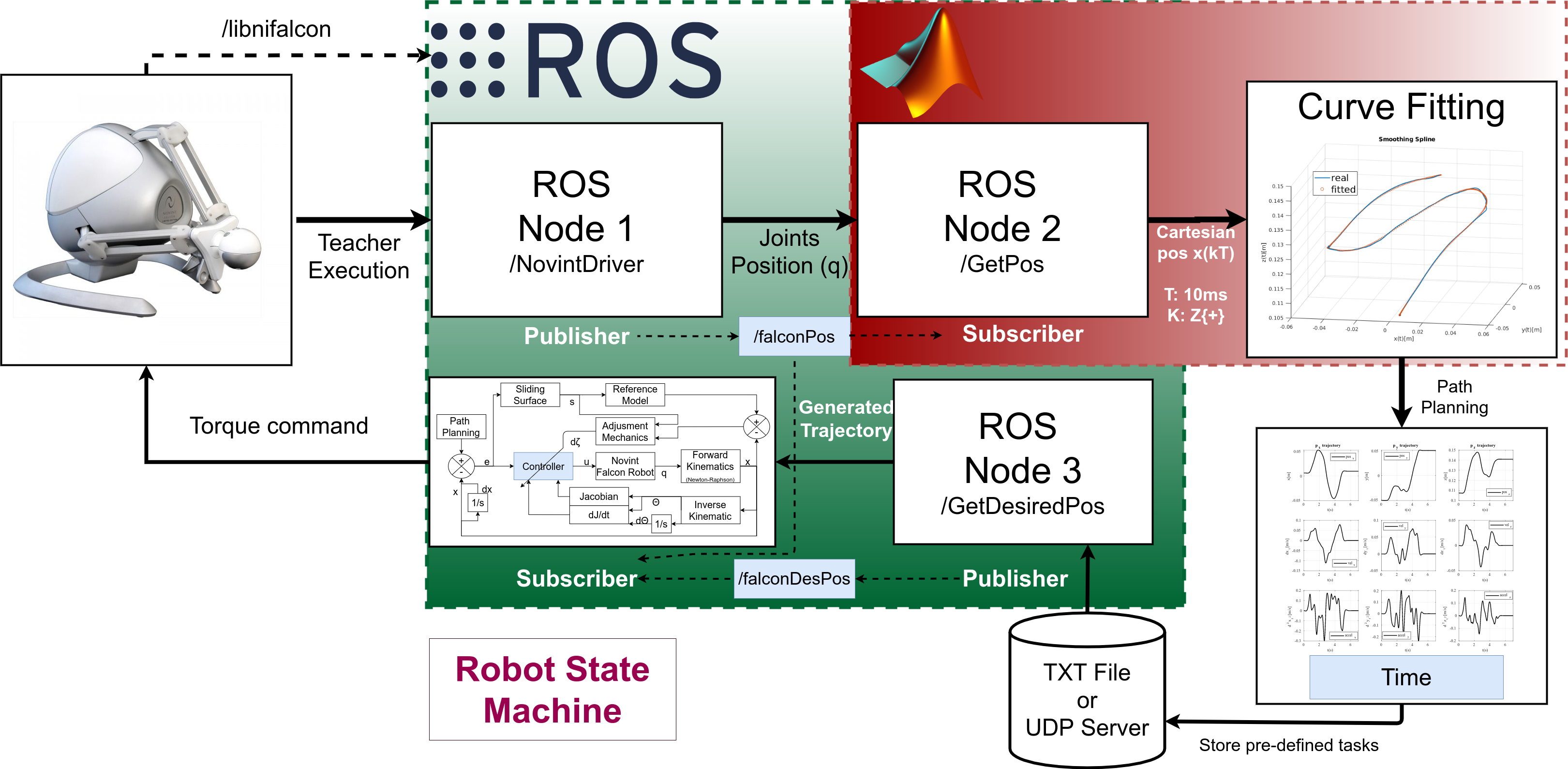}}
\caption{Communication diagram between a user and the Novint Falcon}
\label{interface}
\end{figure*}

We implemented a low-cost device with the idea that a system could potentially be used in homes. The Novint Falcon haptic device has a mechanism with three rotational actuators mounted to a base, and a series of kinematic parallelogram legs to constrain the motion of the end-effector to only translational movements. This form has proven itself as an excellent platform for safe pick-and-place operations due to the mechanism’s low actuated inertia when compared to serial counterparts \cite{b9}. In \cite{b10} the Falcon’s workspace is defined as a bounded tri-hemispherical region overlapping along the common longitudinal (z) axis.

In order to obtain motion kinematics from the trajectories performed as they are made, an interface shown in Fig.~\ref{interface} between ROS and Matlab was developed. Through a USB interface, connected to the Novint, one can send the actuation commands and read encoders data via the interface. Although Novint has released a closed-source SDK, for this research, in order to issue torque commands directly to the motor, the open-source driver “libnifalcon” is employed.
The received encoder values are related to the angular positions $(q_1, q_2, q_3)$. Using forward kinematics we can obtain the Cartesian points of the end-effector.

In demonstration mode, a user performs end-effector movements. Through the interface, we get points of the path followed by the end-effector. These points, along with their sampling time, are used to fit a curve by smoothing splines.  The obtained curve is derived by time to obtain the velocity and acceleration of the end-effector in Cartesian space. The generated trajectory, along with their sampling time, $(q, \dot{q},\ddot{q}, t_{s})$ is stored for later use in reproduction mode with different toys.

\section{Falcon Robot Model}
\subsection{Kinematic Modelling}
The manipulator kinematics consists of a mapping between the joints angle and the end-effector Cartesian position. The Novint Falcon has a unique solution for inverse kinematics, however, the forward kinematic solution is not straightforward. To overcome this issue, an iterative procedure based on the Newton-Raphson method was employed. For the sake of brevity, interested readers are referred to \cite{b11} for the detailed information about the equations.

The inverse kinematics operation results in three joint angles, $\theta_{1i},\theta_{2i},\theta_{3i}$, for each leg $i$, which can define a possible configuration of each leg for the given position of the end-effector. 
Finally, the differential kinematics is defined by
\begin{equation}
\dot{\vec{q}} = J\dot{\vec{x}}
\label{eq1}
\end{equation}

where $\dot{\vec{q}} = [\dot{\theta}_{11}, \dot{\theta}_{12}, \dot{\theta}_{13}]$ represents a set of actuated joint angles, $J$ is the Jacobian matrix and $\dot{\vec{x}}$ is a 3-dimensional velocity vector of the end-effector.
It will be shown in upcoming sections that the time differentiation of the Jacobian matrix, $\dot{J}$, will be needed in some equations.  As defined in \cite{b11}, the Jacobian matrix $J$ is a function of $\theta_{ij}$, which are functions of time. Consequently, $\dot{J}$ will be a function of $\theta_{ij}$ and $\dot{\theta}_{ij}$. It should be noted that $\dot{\theta}_{ij}$ is obtained by numerical differentiation of $\theta_{ij}$, and this approach inherently has estimation errors.

\subsection{Dynamic Modelling}
The dynamic system of the Novint Falcon is described in detail in \cite{b10}, where the experimental identification is made using the open-source “libnifalcon”. The dynamic model is described as
\begin{equation}
\begin{aligned}[b]
\tau ={} &c_m
    \begin{bmatrix}
        sin(\phi_{1})sin(\vec{q_{1}}+\varphi) \\
        sin(\phi_{2})sin(\vec{q_{2}}+\varphi) \\
        sin(\phi_{3})sin(\vec{q_{3}}+\varphi)
    \end{bmatrix}
    +c_{I}(\ddot{\vec{q}})+c_{s}sign(\dot{\vec{q}})\\
    &+c_{d}(\dot{\vec{q}})+(J)^{-T}(m(\vec{a_{ p}}+\vec{g})+F_{ext})
\end{aligned}\label{eq2}
\end{equation}

\noindent where, \\
\indent$\tau$ is the torque applied to the actuators; \\
\indent$c_{m}, c_{i}$ are constants related to the mass of the links;\\
\indent$m$ is a constant related to the mass of the end-effector;\\
\indent$c_{d}$ is the viscous damping of the actuator;\\
\indent$c_{s}$ is the dry friction coefficient of the actuator;\\
\indent$\varphi$ Is the offset angle of input link center of mass;\\
\indent$\vec{a}_{p}$ is the acceleration of the end effector;\\
\indent$F_{ext}$ is the external force applied to the end-effector.\\

In order to obtain $\vec{a}_p$, one can perform a differentiation with respect to time of the differential kinematics equation (i.e., \eqref{eq1}) to obtain the equation
\begin{equation}
    \dot{\vec{a}}_{p} = J^{-1}\ddot{q} - J^{-1}\dot{J}J^{-1}\dot{q}
\label{eq3}
\end{equation}

\subsection{Kinematic and Dynamic Parameters}
The corresponding kinematic and dynamic parameters are presented in Table \ref{tab1}. It is important to clarify that the estimated parameters in \cite{b10} are not in SI base units, since the torque commands that are sent to the robot firmware are also not defined in SI base units. However, it does not modify the accuracy of the applied control algorithm.

\begin{table}[htbp]
\caption{Estimated parameters of the Novint Falcon}
\begin{center}
\begin{tabular}{|c|c|} 
\hline
Parameter & Value \\ [0.5ex] 
\hline\hline
$c_{m}$ & -192 \\ 
\hline
$c_{I}$ & 5.5 \\ 
\hline
$c_{d}$ & 33 \\ 
\hline
$c_{s}$ & 112 \\ 
\hline
$m$ & 486 \\ 
\hline
${\varphi}$ & $\frac{\pi}{6}$ \\ 
\hline
$\vec{\phi}$ & $[\frac{7\pi}{12}, \frac{-pi}{12}, \frac{-9pi}{12}]^{T}$ \\ 
\hline
$\vec{g}$ & $[0, g, 0]^{T}$\\ 
\hline
\end{tabular}
\label{tab1}
\end{center}
\end{table}
\section{Adaptive Tracking Control}
\subsection{Controller Structure}
For control purposes, the dynamic model is expressed in Cartesian coordinates based on \eqref{eq1}, \eqref{eq2} and \eqref{eq3}. We get
\begin{equation}
    J^{T}\tau = M_{x}(q)\ddot{x} + S_{x}(q,\dot{q})\dot{x} + D_{x}(\dot{q}) + G_{x}(q) + F_{ext}
\label{dynamics}
\end{equation}

\begin{figure}[htbp]
\centerline{\includegraphics[scale=0.17]{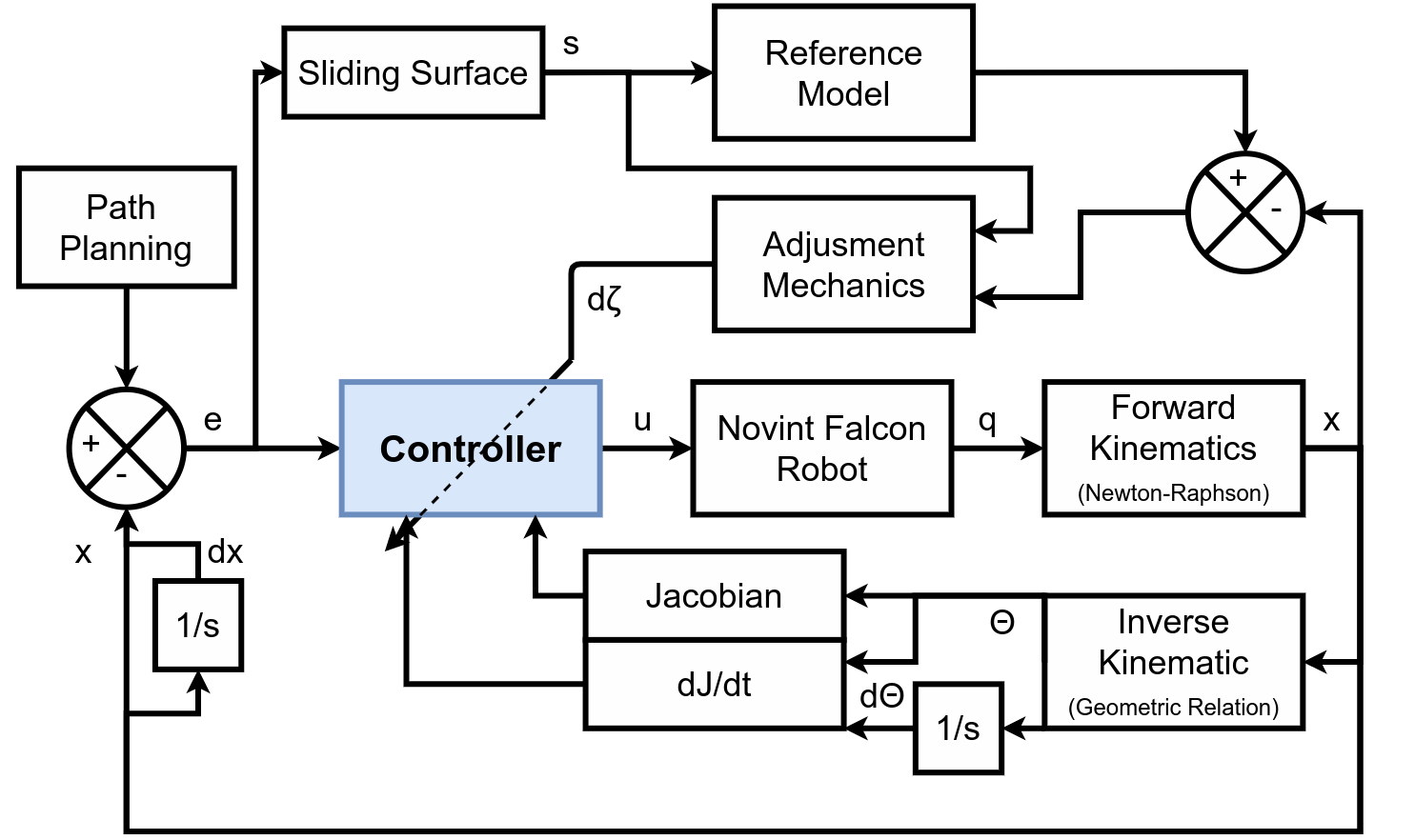}}
\caption{Adaptive tracking control scheme}
\label{fig:controlScheme}
\end{figure}

\noindent where,
\begin{subequations}
\begin{align}
    M_x(q) &= J^Tc_IJ+Im \label{dyn:a}\\
    S_{x}(q,\dot{q}) &= J^{t}c_{I}\dot{J} + J^{T}c_{d}J \label{dyn:b}\\
    D_{x}(\dot{q}) &= J^{T}c_{s}sign(\dot{q}) \label{dyn:c}\\
    G_x(q) &= J^Tc_m<\sin(\vec{\phi}),\sin(\vec{q}+\varphi)>+m\vec{g} \label{dyn:d}
\end{align}  
\end{subequations}
Inspired in \cite{b8} and \cite{b9}, we define an error vector $s = \dot{e} + {\Lambda}e$, where $e=x-x_d$ represents the unconstrained path-tracking error, with $x_d \in {\mathbb{R}}^{3}$ being the desired trajectory; and $\Lambda=diag(\lambda_1,\lambda_2,\lambda_3)$ a diagonal positive definite matrix of tuning parameters, with $\lambda_i>0$. We get
\begin{equation}
    \dot{x} = s + \dot{x}_d - \Lambda e
\label{dyn:sa}
\end{equation}

If there is no robot-environment interaction (e.g., $F_{ext}=0$), the dynamic model of the robot, defined in equation \eqref{dynamics}, is rewritten as
\begin{equation}
    J^{T}\tau = M_{x}(\dot{s} + \ddot{x}_{d} - {\Lambda}\dot{e}) + S_{x}(s+\dot{x}_{m}-{\Lambda}e) + G_{x} + D_{x}
\label{dyn:control}
\end{equation}

It should be noted that while grasping an object, the mass of the end-effector will change. This implies that $m$ is not known and should be estimated, and to achieve that a matrix $\Gamma$ is defined as
\begin{equation}
    \Gamma=I\gamma\overline{m}
\label{Gamma}
\end{equation}

where $\Gamma\in\mathbb{R}^{3x3}$ is a symmetric positive definite Matrix; $\gamma\in[0,1]$ is a tuning parameter and $\overline{m}$ is the maximum value that $m$ can have. 

\subsection{Controller Design}
The adaptive tracking control scheme, shown in Fig. \ref{fig:controlScheme}, was based on the concept of inverse dynamics (e.g., \cite{b5}, \cite{b8}), and is defined as
\begin{equation}
    \tau = J^{T}[\hat{M}_{x}(\ddot{x}_{d}-{\Lambda}\dot{e}) + S_{x}(\dot{x}_{d} - {\Lambda}e) + \hat{G}_{x} + D_{x} - K_{d}s]
\label{control:apadt}
\end{equation}

\noindent where, 
\begin{equation}
\begin{aligned}[b]
    \hat{M}_{x} &= J^{T}c_{I}J+\hat{\Gamma}, \\
    \hat{G}_x &= J^Tc_m<\sin(\vec{\phi}),\sin(\vec{q}+\varphi)>+\Gamma\vec{g}
\end{aligned}
\label{gamma:applied}
\end{equation}
represents the estimate of $M_x,G_x$ respectively. The estimate of $\gamma$ is represented in $\widehat{\Gamma}$ by $\Gamma=I\overline{m}\widehat{\gamma}$. 
Observe that by combining \eqref{dyn:control} and \eqref{control:apadt} the following dynamical system is obtained:
\begin{equation}
\begin{aligned}[b]
    M_{x}\dot{s} + (M_{x} - \hat{M}_{x})[\ddot{x}_{d} - {\Lambda}\dot{e}] +S_{x}s + (G - \hat{G}) + K_{d}s  &= 0, \\
    M_{x}\dot{s} + (\Gamma - \hat{\Gamma})[\ddot{x}_{d} - {\Lambda}\dot{e}] +S_{x}s + (\Gamma - \hat{\Gamma})\vec{g} + K_{d}s  &= 0
\end{aligned}
\label{eq:10}
\end{equation}
where $(\Gamma - \hat{\Gamma}) = I\overline{m} \tilde{\gamma}$ and $\tilde{\gamma} = \gamma - \hat{\gamma}$. If we design and appropriate update law such that $\hat{\Gamma} \to \Gamma$, then \eqref{eq:10} becomes
\begin{equation}
    M_{x}\dot{s} + S_{x}s + K_{d}s  = 0
\end{equation}

Now, in order to proceed with the stability analysis, the following Lyapunov function is defined by
\begin{equation}
    V(s,\tilde{\gamma}) = \frac{1}{2}s^{T}M_{x}s+\frac{1}{2}\tilde{\gamma}^{2}k
\label{lyapunov}
\end{equation}
where k is a positive constant related to the adaptive law. Since $M_{x}$ is a positive definite matrix, $V(s,\tilde{\gamma})$ is, likewise, positive definite. Additionally, notice that $s^TM_ss\to\infty$ as ${\lVert}s{\rVert} \to \infty$ and $\tilde{\gamma}^{2}k \to \infty$ as ${\lVert}\tilde{\gamma}{\rVert} \to \infty$, which proves that $V(s,\tilde{\gamma})$ is radially bounded.

The time derivative of $V(s,\tilde{\gamma})$ along the system trajectory, described in \eqref{lyapunov}, can be computed as
\begin{equation}
    \dot{V}(s,\tilde{\gamma}) = -s^{T}\tilde{\gamma}[\overline{m}(\ddot{x}_{d}-{\Lambda}\dot{e})]-s^{T}[k_{d}+J^{T}c_{d}J]s+k\tilde{\gamma}\dot{\tilde{\gamma}}
\label{lyapunov:2}
\end{equation}
where $\tilde{\gamma}$ is replaced by $-\hat{\gamma}$. Therefore, under the assumption,
\begin{equation}
    \dot{\hat{\gamma}} = -\frac{1}{k}[s^{T}(\overline{m}(\ddot{x}_{d}) + \vec{g})]
\label{adaptive_law}
\end{equation}
\noindent equation (\ref{lyapunov:2}) becomes,
\begin{equation}
    \dot{V}(s,\tilde{\gamma})=-s^{T}[k_d+J^{T}c_{d}J]s
\label{lyap2:final}
\end{equation}

Therefore ($s, \hat{\gamma}$) are uniformly bounded and $V(s,\gamma){\le}0{\ }{\forall}(s,\gamma)\in\mathbb{R}^{3}{\times}{\mathbb{R}}^{1}$ with $\dot{V}(s,\gamma) = 0 {\Leftrightarrow}(s,\gamma)=(0_{3},0)$. Hence, we can conclude that $s \in \mathcal{L}_{\infty}\cap\mathcal{L}_{2}$. From \eqref{eq:10} we conclude that $\dot{s}$ is bounded; as a result, according to the Barbarat’s lemma, the asymptotic convergence of $s$ can easily be proven. This further implies that the error $e =x-x_{d}$ converges, which completes the proof.
\section{Procedure}
The procedure was as follows:\\
- We performed a trajectory from the point (0.005; -0.052; 0.108) to the point (0; 0:05; 0:14) without supporting a toy. The interface tracked this trajectory, and extracted information such as position, velocity, etc.\\
- The controller had as input the demonstrated trajectory information (velocity, position and acceleration), and followed this trajectory. The parameters of the proposed adaptive tracking controller scheme (e.g., \eqref{control:apadt} and \eqref{adaptive_law}) were selected as $K_d = 20 I_3, k = 500, \hat{\gamma}_0=0.5$, where $\hat{\gamma}_0$ is the initial parameter of $\hat{\gamma}$.\\
- In the reproduction phase, after approximately 1.5 seconds a toy was added to the end-effector of the robot, and taken off at approximately 5 seconds. The controller estimated this variation, and modified its internal parameters, and thus, the use of an adaptive controller is supported.\\
- The error between the demonstrated trajectory (without a load variation) and the reproduced trajectory (that supported the load) was calculated. Finally, the parameter $\lambda$, which is related to the end-effector load (as seen in equation \eqref{Gamma}), was estimated. 

\section{Results}
\begin{figure}[htbp]
\centerline{\includegraphics[scale=0.19]{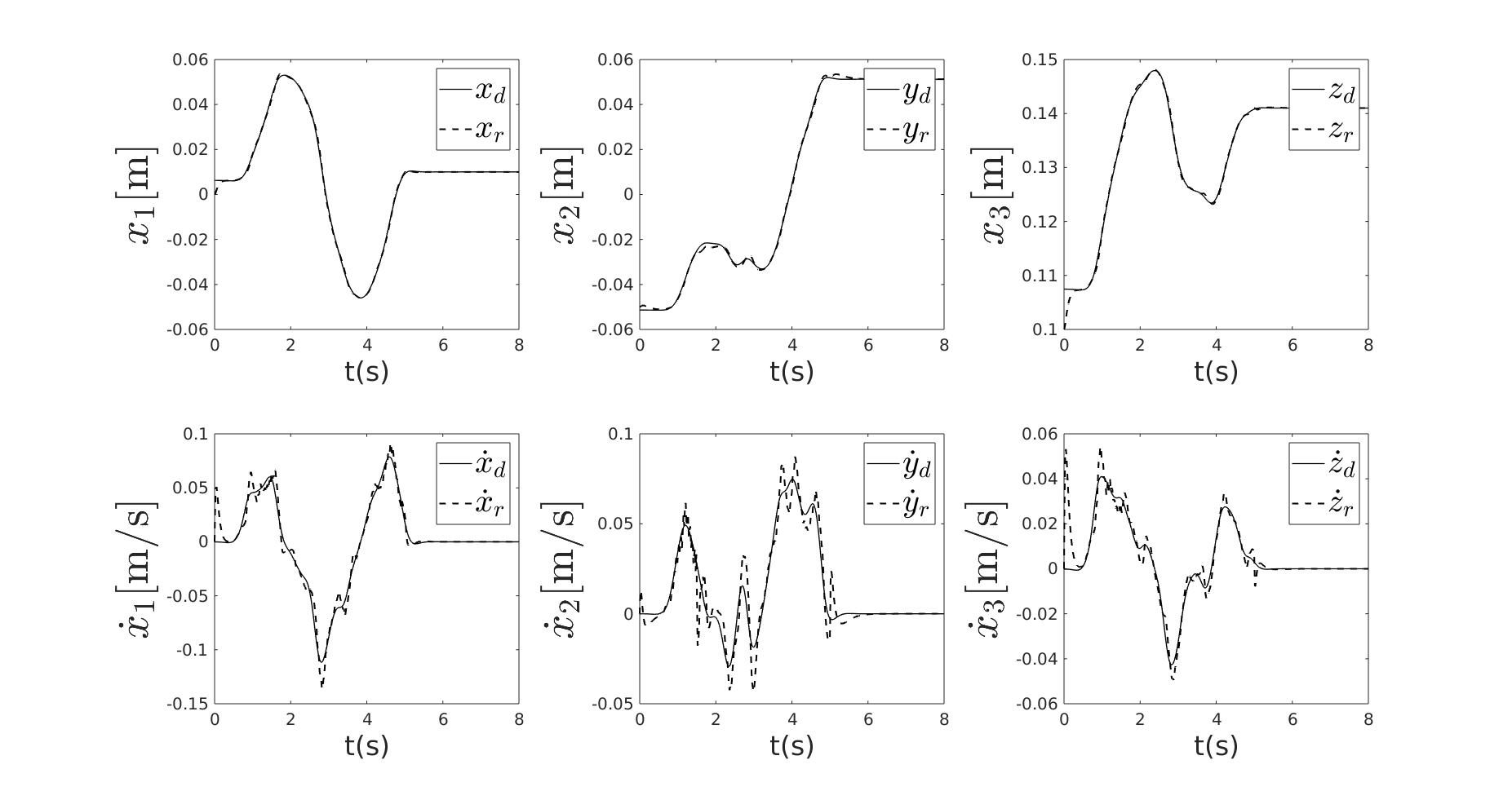}}
\caption{Comparison between demonstrated (solid line) and reproduced (dotted line) position and velocity of the end-effector in Cartesian x (left panel), y (middle panel) and z (right panel) coordinates}
\label{fig:pos_vel}
\end{figure}

\begin{figure}[htbp]
\centerline{\includegraphics[scale=0.19]{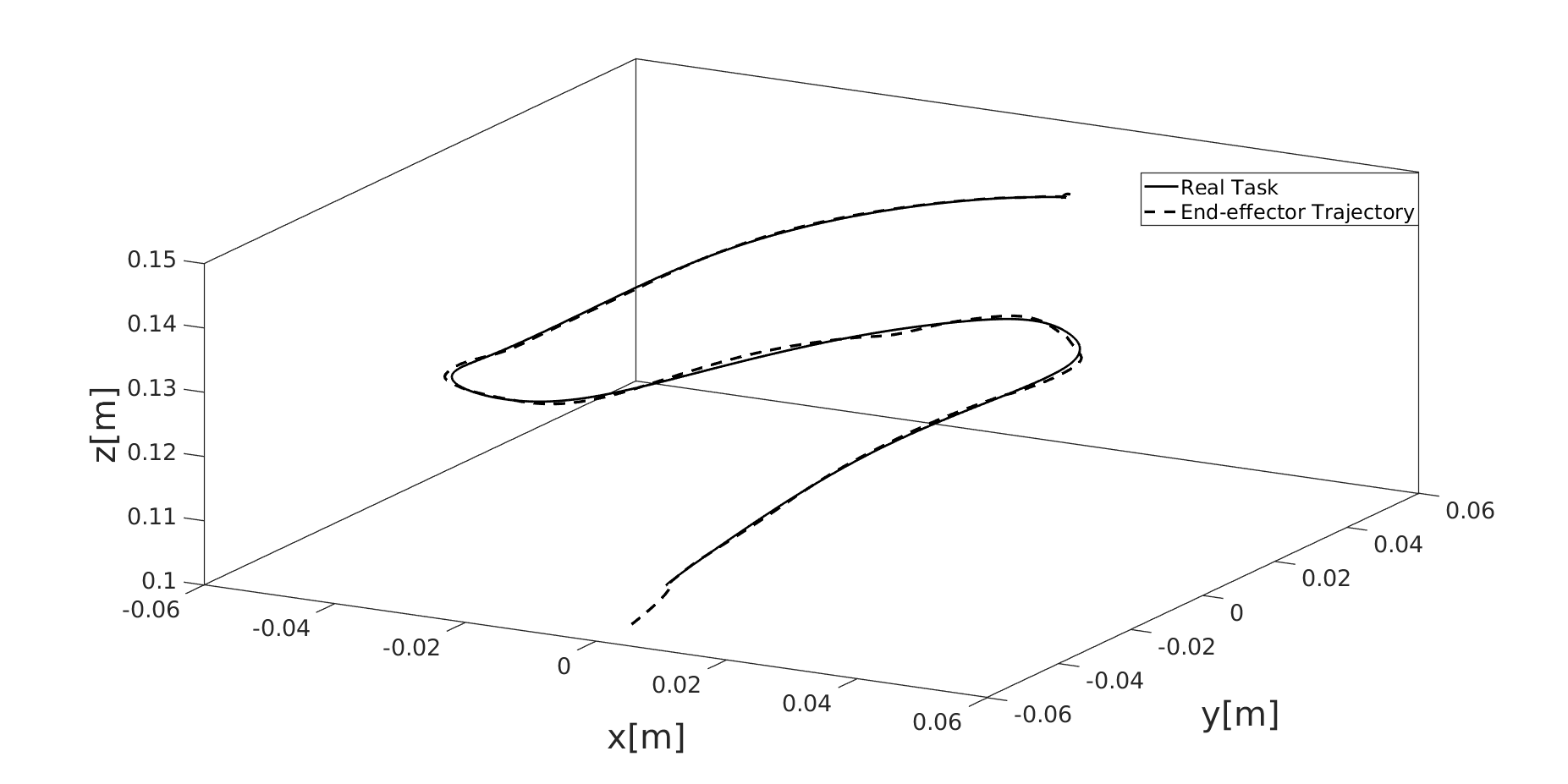}}
\caption{Cartesian 3D plot of the end-effector movement in the demonstrated (solid line) and reproduced (dotted line) trajectories}
\label{fig:path}
\end{figure}

The position and velocity vectors ($\vec{x}(t), \dot{\vec{x}}(t)$), where $\vec{x}(t)$ was expressed as $(x, y, z)$, related to the performed tasks, were stored as a function of time as shown in Fig. \ref{fig:pos_vel}.\\
Figure \ref{fig:path} shows the demonstrated and reproduced end-effector trajectories in 3 dimensions. It is possible to observe that the demonstrated trajectory is accurately reproduced by the end-effector even when supporting the toy.

\begin{figure}[htbp]
\centerline{\includegraphics[scale=0.19]{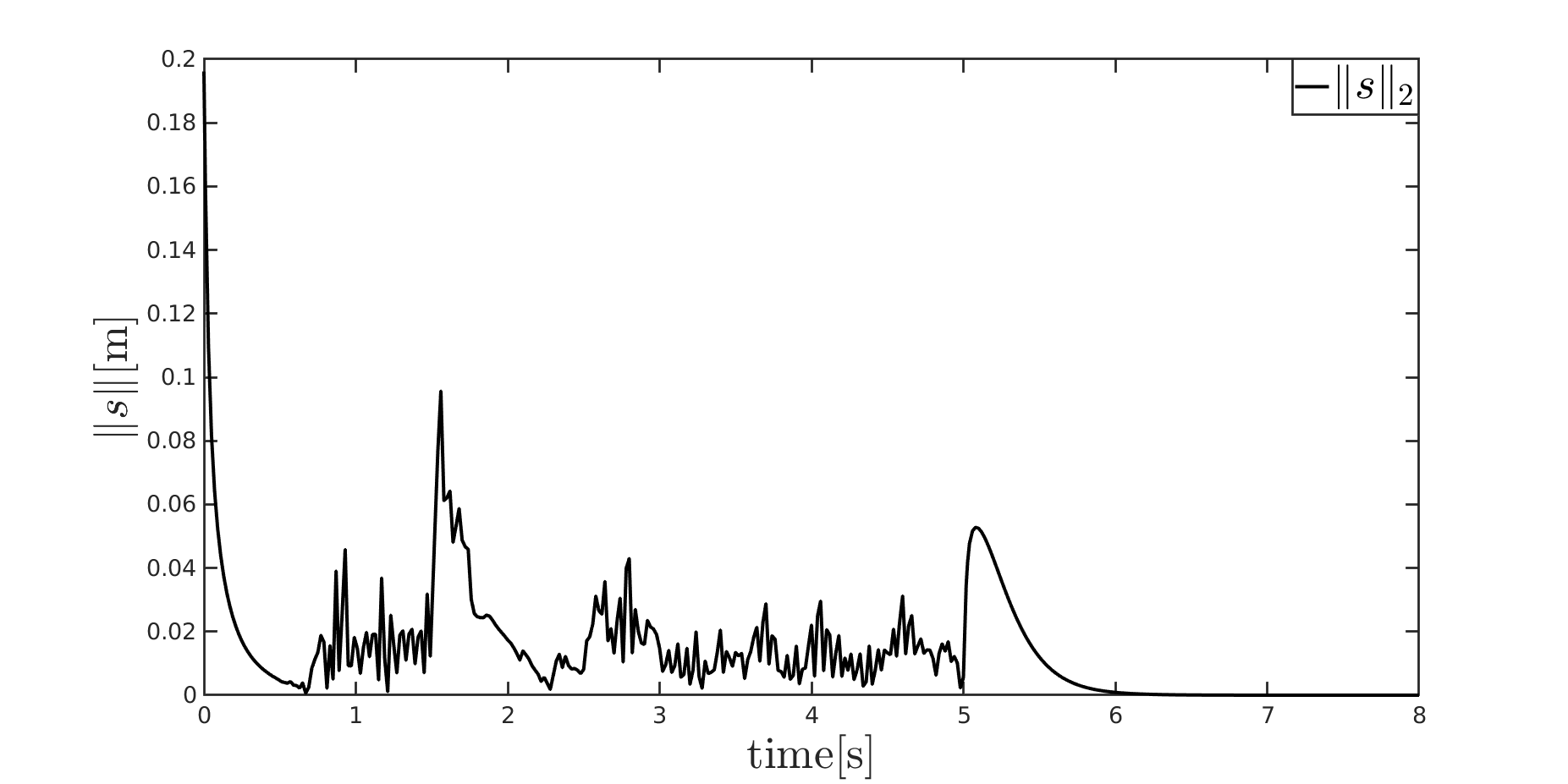}}
\caption{Time evolution of error module $\left\lVert s\right\rVert$}
\label{fig:error}
\end{figure}

\begin{figure}[htbp]
\centerline{\includegraphics[scale=0.19]{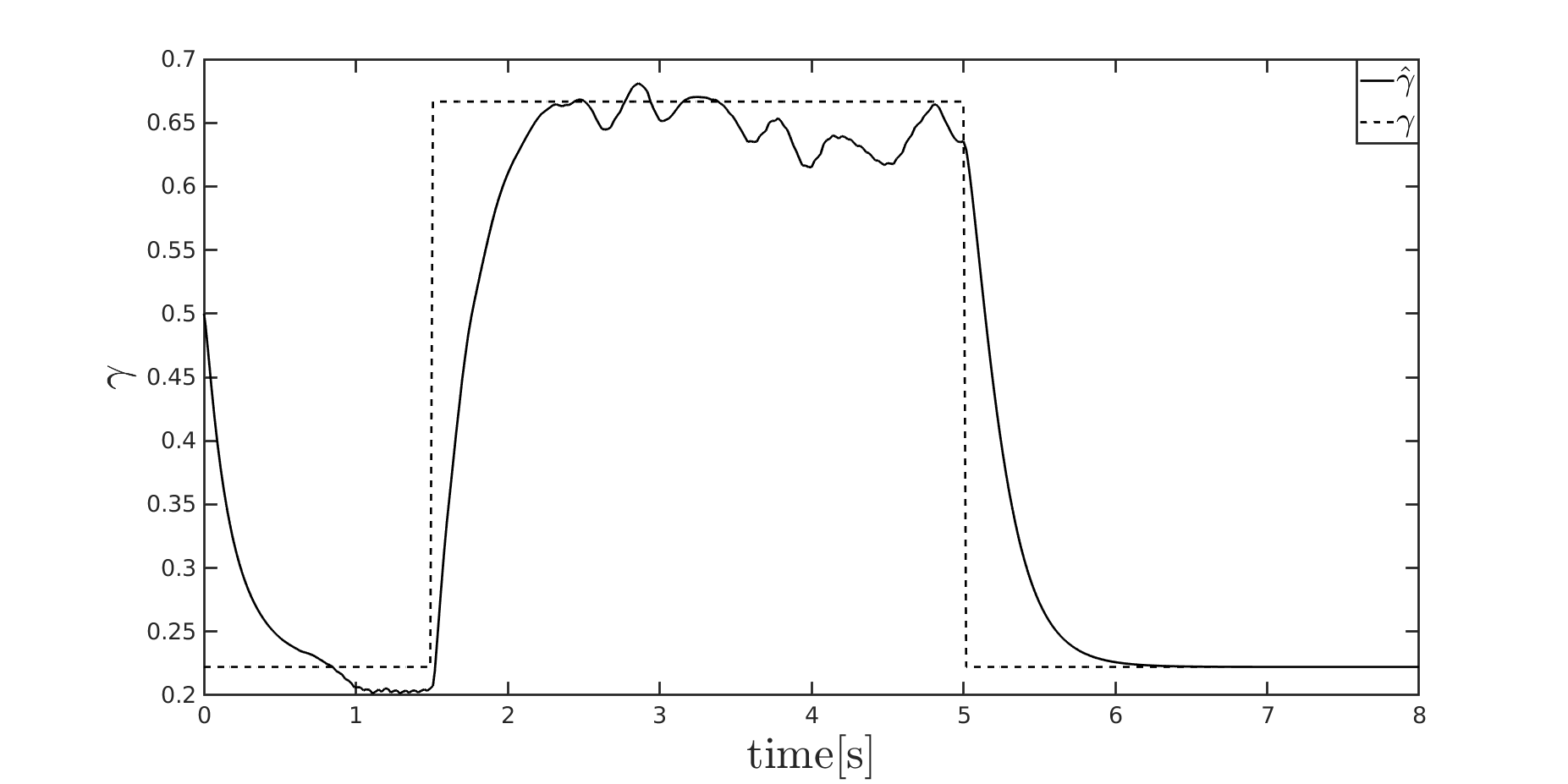}}
\caption{Time evolution of parameter $\lambda$}
\label{fig:gamma}
\end{figure}

The time evolution of the error module $\left\lVert s\right\rVert$ is shown in Fig. \ref{fig:error}. It can be seen that the error increased slightly when the toy is added (1.5 s) and removed (5 s) but settles to minimal quickly. The time evolution of the parameter $\lambda$, increases and decreases related to the end-effector load (i.e. the small toy), as shown in Fig. \ref{fig:gamma}.

Therefore, as can be seen in Fig. \ref{fig:path} and Fig. \ref{fig:error}, the reproduced trajectory follows the demonstrated path very well, as well as the load variation is accurately estimated by the adaptive controller, as shown in Fig. \ref{fig:gamma}.

\section{Conclusion and Future Works}
In this paper, we have proposed a robust Novint Falcon controller that can adapt to different trajectories and payload weights. In this way it can support spontaneous play tasks and different toys for children with disabilities. The suitable performance of the control scheme is supported by the stability analysis in the Lyapunov sense.

For its practical implementation, a numerical differentiation of variables (i.e. $\vec{q}, J$) are required, which will inherently generate error in the controller's performance, as can be seen in Fig. \ref{fig:pos_vel}, where the velocity is not tracked as well as the position was. In order to avoid this issue, a set of gyroscopes, which gives an accurate measurement of the angular velocities, should be used by placing them on the actuator's legs.

Future work will support inputs such as brain computer interfaces and single switches so that the user can instigate the reproduction of the trajectory independently. In addition, we will develop an impedance adaptive controller to reproduce smooth movements over a trajectory. After implementing these steps, the system can be tried by children to enhance their play experience and safety.

\section*{Acknowledgment}
This research was supported by a Collaborative Health Research Project (CHRP), a joint initiative of the National Sciences and Engineering Research Council (NSERC) and Canadian Institutes of Health Research (CIHR), and University of Alberta Research Experience funding.

\end{document}